\documentclass[11pt,a4paper]{article}
\usepackage[hyperref]{acl2019}

\usepackage{times}
\usepackage{latexsym}
\usepackage{graphicx}
\graphicspath{ {./images/} }

\usepackage{url}
\usepackage{algorithm,algpseudocode}
\usepackage{amsmath}
\usepackage{amsfonts}
\usepackage{amssymb}

\title{No More Distractions: an Adaptive Up-Sampling Algorithm to Reduce Data Artifacts}

\author{First Author \\
  Affiliation / Address line 1 \\
  Affiliation / Address line 2 \\
  Affiliation / Address line 3 \\
  \texttt{email@domain} \\\And
  Second Author \\
  Affiliation / Address line 1 \\
  Affiliation / Address line 2 \\
  Affiliation / Address line 3 \\
  \texttt{email@domain} \\}

\date{}

\begin{document}

\maketitle
\begin{abstract}
  Researchers recently found out that sometimes language models achieve high accuracy on benchmark data set, but they can not generalize very well with even little changes to the original data set. This is sometimes due to data artifacts, model is learning the spurious correlation between tokens and labels, instead of the semantics and logic. In this work, we analyzed SNLI data and visualized such spurious correlations. We proposed an adaptive up-sampling algorithm to correct the data artifacts, which is simple and effective, and does not need human edits or annotation. We did an experiment applying the algorithm to fix the data artifacts in SNLI data and the model trained with corrected data performed significantly better than the model trained with raw SNLI data, overall, as well as on the subset we corrected.
\end{abstract}

% \section{Credits}

% This document has been adapted from the instructions

% instructions of the \emph{International Joint Conference on Artificial
%   Intelligence} and the \emph{Conference on Computer Vision and
%   Pattern Recognition}.
\section{Introduction}
Natural language models are expected to learn and understand the semantics of text, so they can "think" like a human and solving tasks such as making inferences and answering questions. However, in recent studies, researchers discovered that sometimes the high performance on benchmark data sets are not a results of the above, instead, they are from learning the spurious correlations between tokens and output label \citep{poliak2018hypothesis}. Spurious correlations are any simple correlations between input tokens and output labels and for a data set without spurious correlations, p(label|token) should be uniform over all class labels \citep{gardner2021competency}. For example, in a language model to predict sentiments of customer reviews. It's possible that 99\% of customers use "perfect" in expressing positive feedbacks, with the remaining 1\% using "perfect" in a sarcasm way to express negative feelings. If such spurious correlation is not eliminated during sampling, it's very likely the correlation will be learned by the model, as the model can easily achieve 99\% accuracy with a simple rule of if "perfect" exists in the input text. This way, model may achieve a very high accuracy in benchmark data sets, but may not be able to generalize very well.

In this study, we will show that such spurious correlations exist in SNLI data \citep{bowman2015large}, in the form that tokens from a specific subset occurs more often with some labels than with other labels. We trained a ELECTRA-small \citep{clark2020electra} model on SNLI data without any correction and observed that these spurious correlations have influenced model training. The model is tend to predict the correlated label if certain tokens exist in input. To solve this issue, the most popular way is manually or semi-automatically edit the records \citep{gardner2021competency,clark-etal-2019-dont,maas-etal-2011-learning}, such that the p(label$|$token) is approximately uniform over all class labels. However, this method will cost significant amount of human time and may introduce additional bias in these augmented data \citep{tafjord-etal-2019-quartz}.

We attempt to remedy this issue by doing an adaptive round robin up-sampling on the records that contains under-represented token-label pair in training data, which has three advantages comparing to the current methods: fully automatic, cost little to none human interactions, not introducing additional bias. With this correction, we observed the accuracy have been improved overall, as well as on the sub set of records with problematic tokens.

\section{Data Artifact Analysis}

The SNLI data was divided into training, validation and test data.
In the training data, after removing punctuation, stopping words and tokenization, for each of the token in the SNLI data premise and hypothesis, we calculated the empirical distribution for its label, and a statistic (equation $\ref{eqn:zstat}$) was constructed to measure the deviation of the distribution from uniform.
\begin{equation} 
\label{eqn:zstat}
p^{*}=max_{l\in L}(\hat{p}_l)
\end{equation}
$p^{*}$ is expected to be close to $\frac{1}{C}$ if there is no spurious correlation, where C is the number of labels and L is the set of labels.

Taking the sample size into consideration, we can construct the following statistics to show how significant the distribution deviate from uniform.
\begin{equation} \label{eqn:zstat2}
z^{*}=max_{l\in L}(\frac{\hat{p}_l-p_l}{\sqrt{p_l(1-p_l)/n}})
\end{equation}
where L is the set of labels, $\hat{p}_l$ is the observed probability of label l given the token, and $p_l$ is the probability of label l given the token under the uniform assumption ($\frac{1}{3}$ for SNLI data).

\begin{figure}[bhp]
\includegraphics[width=\columnwidth]{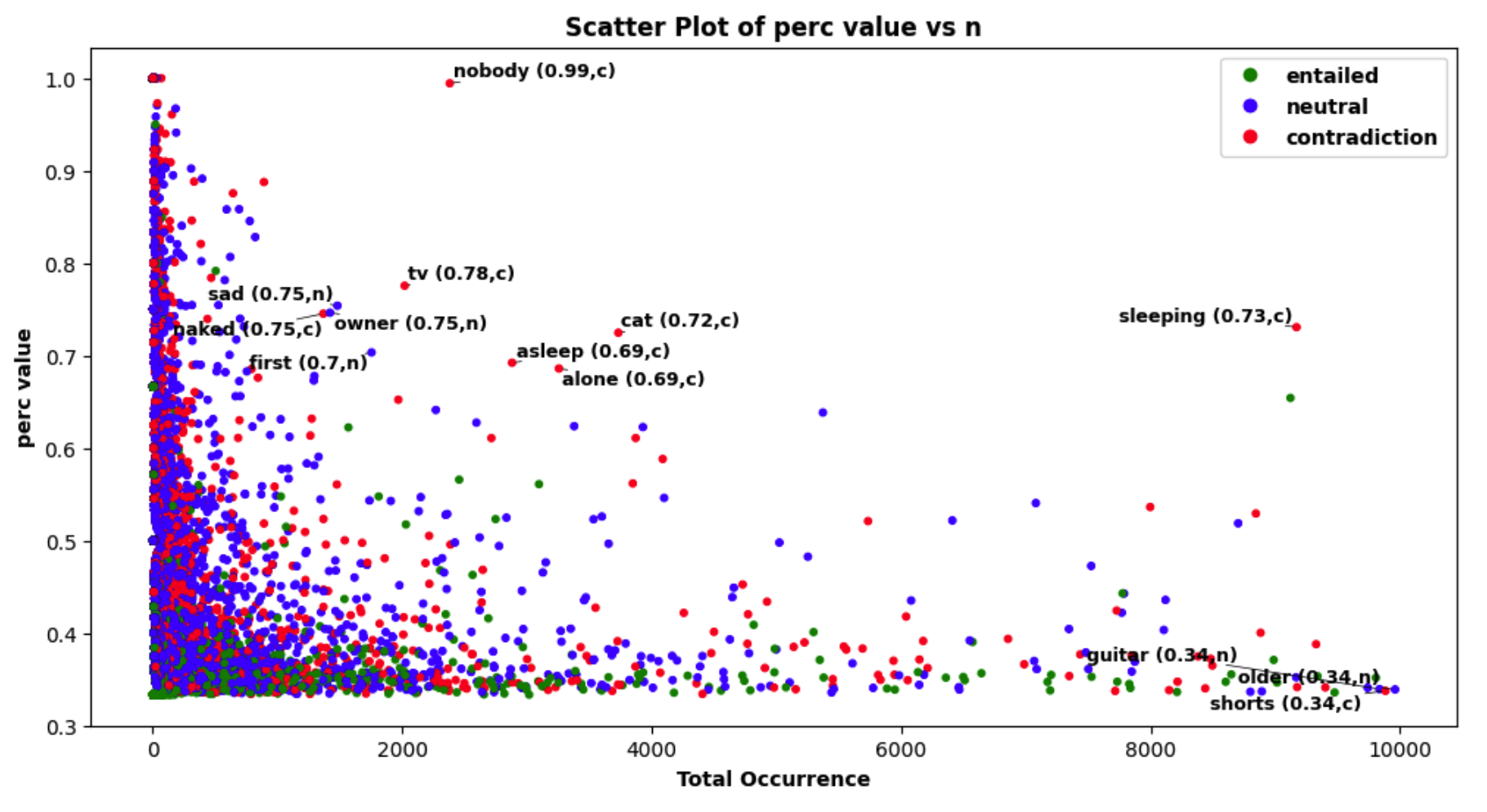}
\includegraphics[width=\columnwidth]{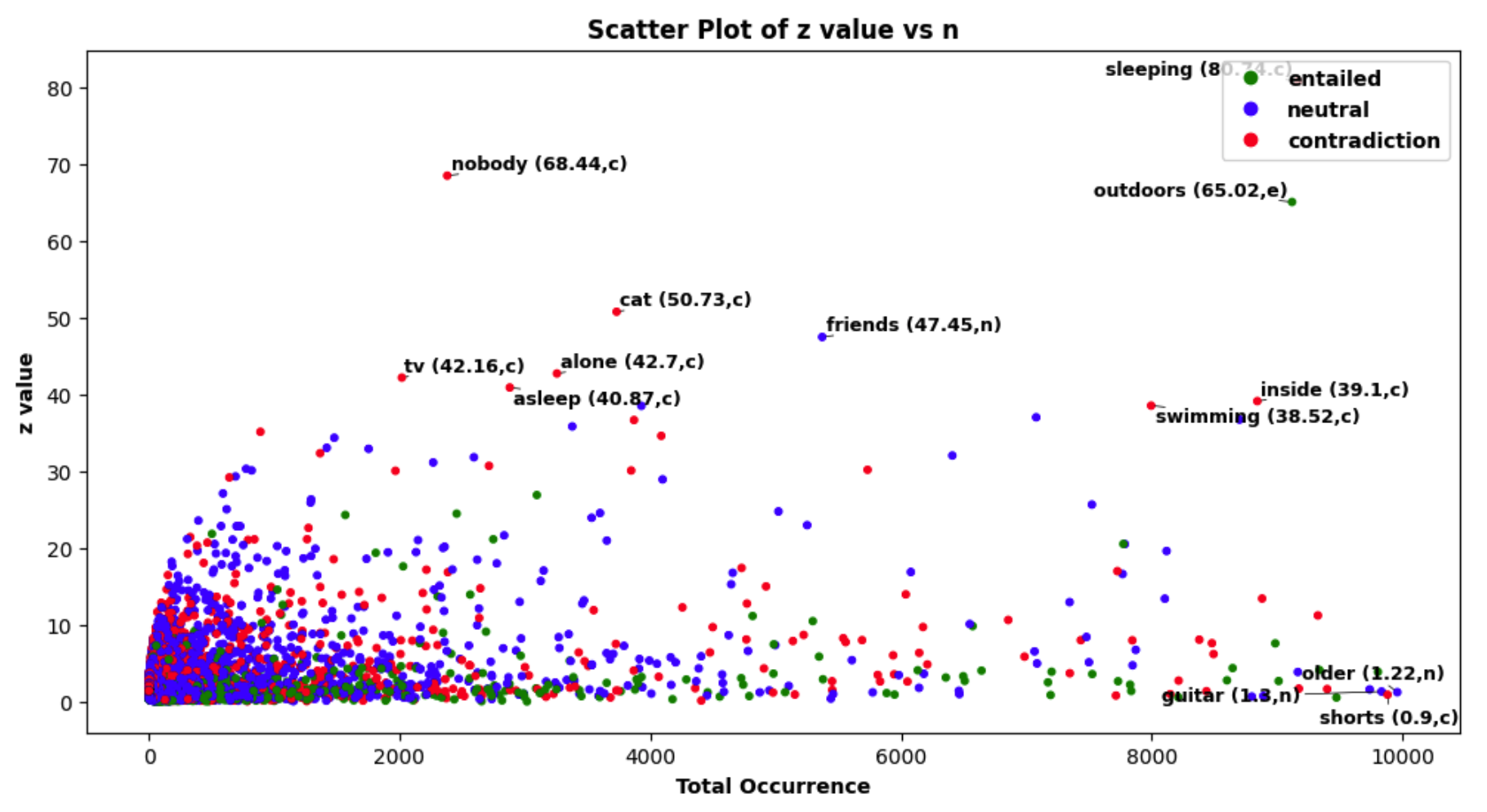}
\caption{Artifact statistics in SNLI}
\medskip
\small
\label{artifacs_before}
Plots of $\hat{p}^*$ and $z^*$ vs the number of occurrences respectively, each dot represent one token. Tokens with the highest $\hat{p}^*$ ($z^*$ in the lower plot) value or number of occurrences are annotated in the format of ($\hat{p}^*$ ($z^*$ in the lower plot) value, label) that associated with the $\hat{p}^*$ ($z^*$ in the lower plot)). Thresholds are applied to show tokens with number of occurrences between 1000 to 10000. 
\end{figure}

We plotted the $\hat{p}^*$ and $z^*$ vs the number of occurrences in Figure \ref{artifacs_before}.
We can see that, tokens like "nobody", "cat", "friends" are very biased towards one class label. Especially for "nobody", 99\% of the time, it occurs together with contradiction label. Use it as an indicator for contradiction, the model can achieve 99\% accuracy on the subset where "nobody" occurs. On a side note, the first plot in Figure \ref{artifacs_before} roughly aligns with the plot in previous research \citep{gardner2021competency}. The difference could be a result of different tokenization method and different thresholds.

\begin{figure}[bhp]

\includegraphics[width=\columnwidth]{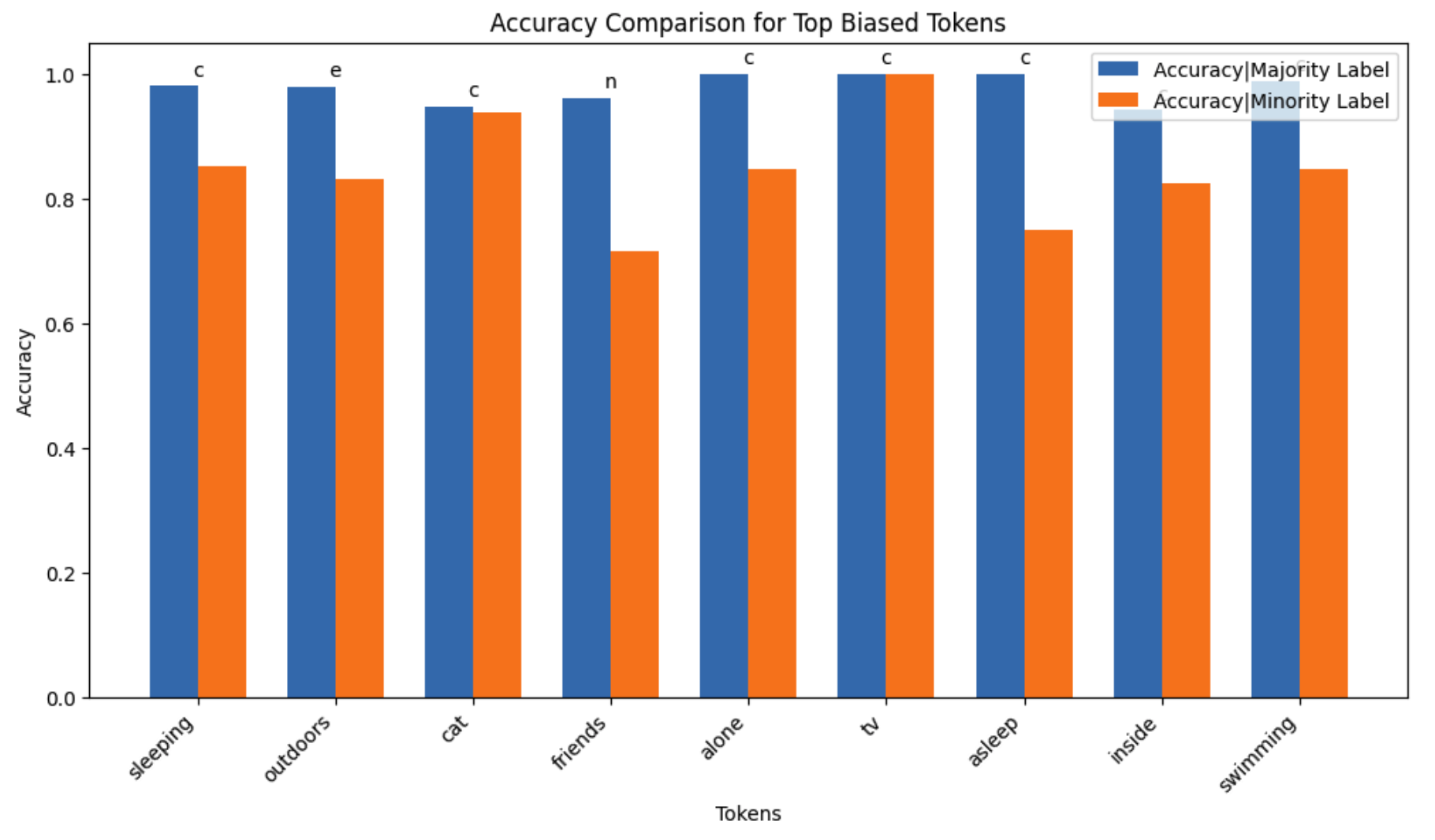}
\caption{Accuracy Comparison for Biased Labels}
\medskip
\small
\label{acc_comp_before}
Each bar is one top biased token (based on $z^*$ value), color represent if the accuracy is for majority label set or minority label set. Blue bars are always higher (or the same) than the red bars.
\end{figure}

We trained and tuned a ELECTRA-small model with training and validation data. In the reserved test data, for each of the top biased tokens (from training data), we collect the records that contain the label, separate them into two sets (based on the distribution of label for that token): majority label set and minority label set. We calculated the accuracy separately for the two set and discovered that model consistently performs better on the majority label set than the minority set. The comparison can be found in Figure \ref{acc_comp_before}.

We examined the cases where top biased tokens occur and predicted label differs from the actual label, which are the errors we are trying to improve in this study. Here are some examples:
\subsection*{Example 1}
\label{error example 1}
premise: Football player jumping to catch the ball with an empty stand behind him.\\
hypothesis: The ball is being thrown the football player direction.\\
true label: entailed\\
predicted label: contradiction\\
involved token: cat
\subsection*{Example 2}
\label{error example 2}
premise: A group of people plays a game on the floor of a living room while a TV plays in the background.\\
hypothesis: A group of friends are playing the xbox while other friends wait for their turn.\\
true label: contradiction\\
predicted label: neutral\\
involved token: friends

In both examples, the model does not understand the meaning and logic of premise and hypothesis, however, the top biased token "cat" (tokenized from "catch") and "friends" do exist in the inputs, it predicts the most popular class label associated with the top biased tokens.
% \begin{table*}[t!]
% \centering
% \begin{tabular}{lll}
%   premise & hypothesis & label & predicted_{label} & token\\
%   \hline
  
%   A reson is in a blue canoe\\ in front of some plants. & a blue canoe inside\\ of some plants	2 &	0 & inside\\
% A man punting a football as fans\\ from the opposing team watch in the\\ background	A man catches a football & 2 & 0 & cat\\
% People working digging a hole.	There are people covering up a hole that someone is stuck inside of. &	2 &	1 &	inside\\
% A baseball player sliding into home as the opposing team's catcher awaits the baseball for the out. & The catcher is holding a bat. &	2 &	1 &	cat\\
% \end{tabular}
% \caption{Citation commands supported by the style file.
%   The citation style is based on the natbib package and
%   supports all natbib citation commands.
%   It also supports commands defined in previous ACL style files
%   for compatibility.
%   }
% \end{table*}
\section{Adaptive Up-Sampling Data Artifacts Correction Algorithm}
From the above analysis, we observed spurious correlations between the tokens in input and the class label exist for benchmark data set like the SNLI data, which may cause the model to learn the spurious correlation, instead of the semantic meaning and logic the model is intended to learn. To deal with this issue, we propose to do an adaptive round-robin up sampling in the training data records that contains the top biased tokens  while carrying class labels that are not the most popular one associated with the token. This up-sampling will bring the $p(label|token)$ to approximately uniform distribution for each token, so that the model will not be "distracted" by learning the spurious correlation in training. The reason we use adaptive round robin is due to some of the records contain multiple tokens which bias toward different class labels, hence we need to adjust the up sampling target in each iteration so that the ending distribution of $p(label|token)$ approximate uniform as close as possible.

The algorithm described above is summarized in Algorithm \ref{alg:1}.

\begin{algorithm}[H]
\caption{Adaptive Up-Sampling Data Artifacts Correction Algorithm (AUDAC)}\label{alg:1}
\begin{algorithmic}[1]
\Require{$T$} \Comment{Training data}
\Require{l, k} \Comment{step size, top k tokens to correct}
\Ensure{$T^*$} \Comment{Corrected Training data}
\Statex

\State $S \gets get\_top\_biased\_tokens(k,T)$
\State $D \gets get\_target\_sample\_size(S,T)$ 
\Comment{dictionary of targeted up-sampling size to achieve uniform}
\State $B \gets get\_toekn\_record\_mapping(S,T)$
\Comment{dictionary of token and index of T (list) mapping}
\While {\texttt{i in range(N) or $D>0$ }} \Comment{early stopping: targeted sample size is close to zero for all tokens}
\For{\texttt{s in S }}
        \State $d = D[s]*l$ \Comment{get sample size for token s}
        \State $T+=random\_sample(B[s],T)$
        \State update D
      \EndFor
\EndWhile
\State $T^*\gets T$

% \State \texttt{}

\end{algorithmic}
\end{algorithm}

\section{Experiments}
To test the performance of the proposed AUDAC algorithm, we applied the algorithm on SNLI training data with k=10 (top 10 biased tokens, due to time and computing power constraint), and step size=0.2. After 10 iterations, the targeted sample size for all biased tokens are close to zero, indicating the distribution are corrected to uniform and the spurious correlation for these ten tokens are eliminated. The training data has increased from 550152 rows to 597580 rows (7.5\% increase).

\begin{figure}[bhp]
\includegraphics[width=\columnwidth]{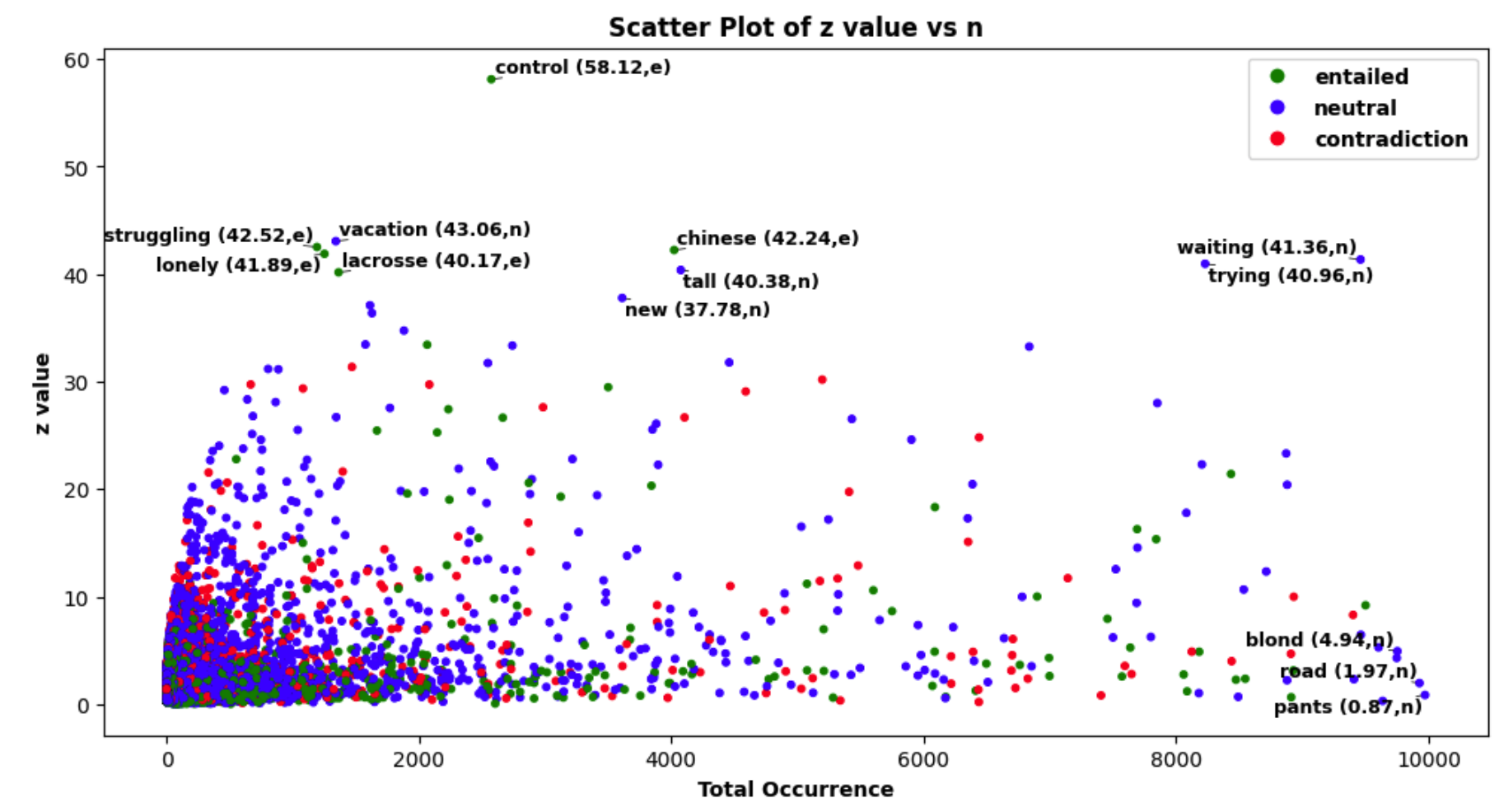}
\caption{Artifact statistics in corrected SNLI}
\medskip
\small
\label{artifacs_after}
\end{figure}

We calculated the the data artifacts statistics ($z^*$) for all tokens with the updated training data, and plotted them against n, similar to Figure \ref{artifacs_before} in Figure \ref{artifacs_after}. Even though spurious correlation still exist in the data, the ten tokens we corrected are no longer the problematic ones. By comparing Figure \ref{artifacs_after} with Figure \ref{artifacs_before}, data artifacts are significantly reduced, with the highest $z^*$ decreasing from 80 (sleeping) to 58 (control).

Now with the improved SNLI training data, we trained a new ELECTRA-small model on it. Then in the test data set we first calculate the overall model accuracy, then separately for the subset of records contain tokens we corrected and the rest, similar to what we did in Section 3.

\begin{figure}[h]
\includegraphics[width=\columnwidth]{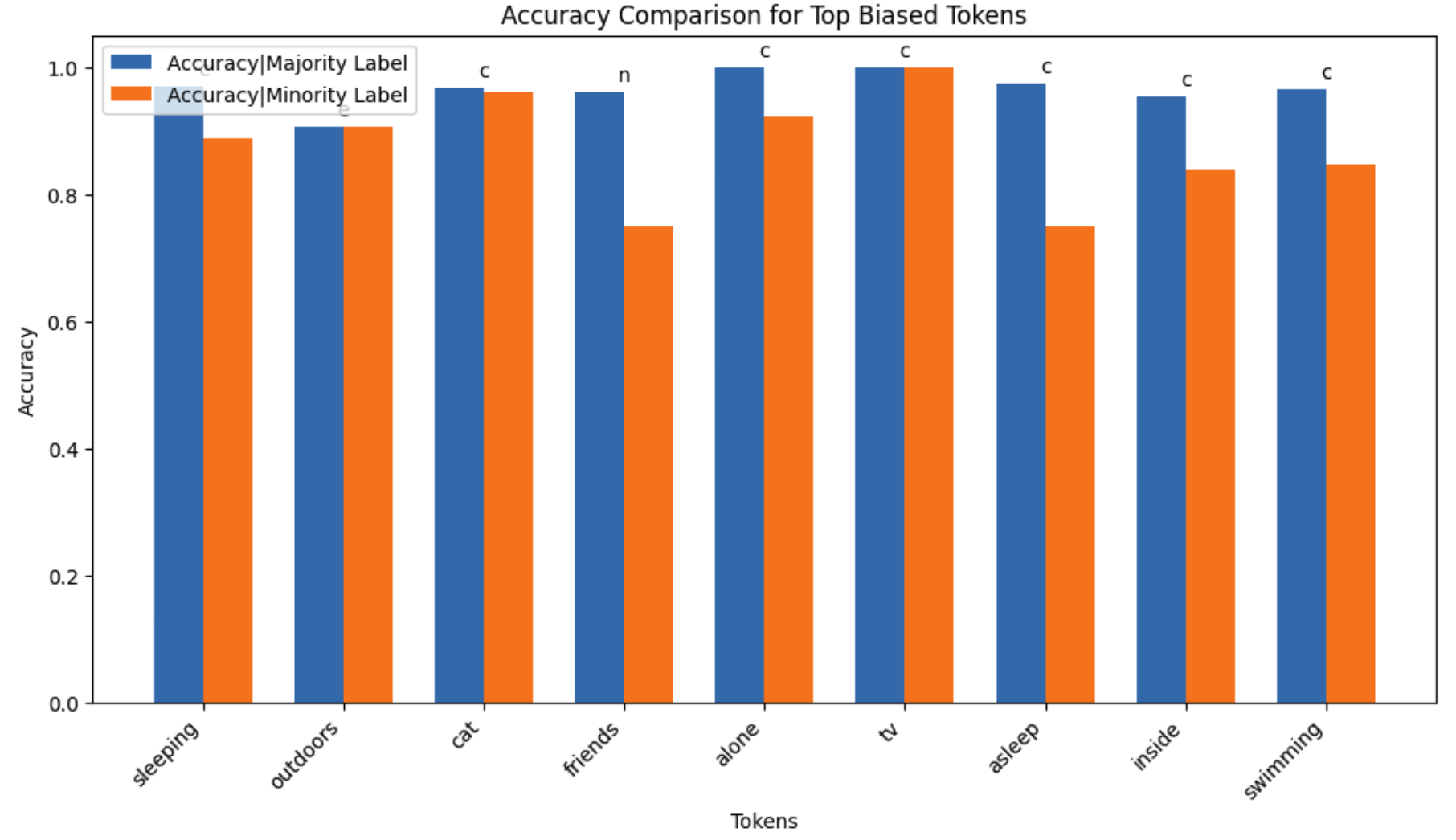}
\caption{Accuracy Comparison for Biased Labels After Correction}
\medskip
\small
\label{acc_comp_after}
Each bar is one top biased token (based on $z^*$ value), color represent if the accuracy is for majority label set or minority label set.
\end{figure}

The overall accuracy has increased from 89.149\% to 89.667\%, which is reasonably good, considering we only corrected ten tokens out of thousands and increased the sample size by 7.5\%. The accuracy on the subset data that contain the tokens corrected increased from 92.94\% to
93.23\%.
Similar to Figure \ref{acc_comp_before}, for each token, we calculated and plotted (Figure \ref{acc_comp_after}) the accuracy for the 
subset of records with the label that the token biased towards to vs the accuracy for the 
subset of records with other labels. 

Between Figure \ref{acc_comp_after} and Figure \ref{acc_comp_before}, we can see that the accuracy for minority label subset has increased quite a bit, closing the gap with the accuracy for majority label subset.

Also in Table \ref{tab:token_accuracy}, we break down the model performance change for each token and whether the record has a majority label or minority label. For most tokens, overall accuracy increased or stayed the same after the correction. The accuracy improved (or stayed the same) for the minority subset associated with all tokens. For majority subset, some of the tokens do have small decrease in accuracy, which was compensated by the increase in minority subset.

\begin{table}[h]
\centering
\small
\scalebox{0.91}{
\begin{tabular}{|c|c|c|c|}
\hline
token & accuracy$|$major & accuracy$|$minor & overall \\
\hline
sleeping & (\textbf{0.98}, 0.97) & (0.85, \textbf{0.89}) & (0.95, 0.95) \\
outdoors & (\textbf{0.98}, 0.91) & (0.83, \textbf{0.91}) & (\textbf{0.93}, 0.91) \\
cat & (0.95, \textbf{0.97}) & (0.94, \textbf{0.96}) & (0.94, \textbf{0.96}) \\
friends & (0.96, 0.96) & (0.71, \textbf{0.75}) & (0.87, \textbf{0.89}) \\
alone & (1.00, 1.00) & (0.85, \textbf{0.92}) & (0.97, \textbf{0.98}) \\
tv & (1.00, 1.00) & (1.00, 1.00) & (1.00, 1.00) \\
asleep & (\textbf{1.00}, 0.97) & (0.75, 0.75) & (\textbf{0.98}, 0.95) \\
inside & (0.94, \textbf{0.95}) & (0.82, \textbf{0.84}) & (0.89, \textbf{0.90}) \\
swimming & (\textbf{0.99}, 0.97) & (0.85, 0.85) & (\textbf{0.93}, 0.92) \\
\hline
\end{tabular}
}
\caption{Accuracy Metrics for Different Tokens}
\label{tab:token_accuracy}
Each value pair represent the accuracy before and after the correction. Bold one is the higher of the two.
\end{table}
\section{Conclusion}
The proposed AUDAC algorithm is simple, cheap (comparing to human edits), yet effective, in correcting data artifacts in training data. We believed that even with the best model architecture, state-of-the-art hardware and software for model training, the model can still be distracted by the spurious correlation or other data artifacts in the training data. With AUDAC algorithm to process the training data, eliminating or reducing data artifacts, then the model will not be distracted by spurious correlation, and focus more on learning to solve the core task in the training process.

\section*{Acknowledgments}

I’d like to thank Dr. Durrett and the TAs for running such a great and helpful NLP class!

\bibliography{acl2019}

\begin{thebibliography}{7}
\expandafter\ifx\csname natexlab\endcsname\relax\def\natexlab#1{#1}\fi

\bibitem[{Bowman et~al.(2015)Bowman, Angeli, Potts, and Manning}]{bowman2015large}
Samuel~R Bowman, Gabor Angeli, Christopher Potts, and Christopher~D Manning. 2015.
\newblock A large annotated corpus for learning natural language inference.
\newblock \emph{arXiv preprint arXiv:1508.05326}.

\bibitem[{Clark et~al.(2019)Clark, Yatskar, and Zettlemoyer}]{clark-etal-2019-dont}
Christopher Clark, Mark Yatskar, and Luke Zettlemoyer. 2019.
\newblock \href {https://doi.org/10.18653/v1/D19-1418} {Don{'}t take the easy way out: Ensemble based methods for avoiding known dataset biases}.
\newblock In \emph{Proceedings of the 2019 Conference on Empirical Methods in Natural Language Processing and the 9th International Joint Conference on Natural Language Processing (EMNLP-IJCNLP)}, pages 4069--4082, Hong Kong, China. Association for Computational Linguistics.

\bibitem[{Clark et~al.(2020)Clark, Luong, Le, and Manning}]{clark2020electra}
Kevin Clark, Minh-Thang Luong, Quoc~V Le, and Christopher~D Manning. 2020.
\newblock Electra: Pre-training text encoders as discriminators rather than generators.
\newblock \emph{arXiv preprint arXiv:2003.10555}.

\bibitem[{Gardner et~al.(2021)Gardner, Merrill, Dodge, Peters, Ross, Singh, and Smith}]{gardner2021competency}
Matt Gardner, William Merrill, Jesse Dodge, Matthew~E Peters, Alexis Ross, Sameer Singh, and Noah~A Smith. 2021.
\newblock Competency problems: On finding and removing artifacts in language data.
\newblock \emph{arXiv preprint arXiv:2104.08646}.

\bibitem[{Maas et~al.(2011)Maas, Daly, Pham, Huang, Ng, and Potts}]{maas-etal-2011-learning}
Andrew~L. Maas, Raymond~E. Daly, Peter~T. Pham, Dan Huang, Andrew~Y. Ng, and Christopher Potts. 2011.
\newblock \href {https://aclanthology.org/P11-1015} {Learning word vectors for sentiment analysis}.
\newblock In \emph{Proceedings of the 49th Annual Meeting of the Association for Computational Linguistics: Human Language Technologies}, pages 142--150, Portland, Oregon, USA. Association for Computational Linguistics.

\bibitem[{Poliak et~al.(2018)Poliak, Naradowsky, Haldar, Rudinger, and Van~Durme}]{poliak2018hypothesis}
Adam Poliak, Jason Naradowsky, Aparajita Haldar, Rachel Rudinger, and Benjamin Van~Durme. 2018.
\newblock Hypothesis only baselines in natural language inference.
\newblock \emph{arXiv preprint arXiv:1805.01042}.

\bibitem[{Tafjord et~al.(2019)Tafjord, Gardner, Lin, and Clark}]{tafjord-etal-2019-quartz}
Oyvind Tafjord, Matt Gardner, Kevin Lin, and Peter Clark. 2019.
\newblock \href {https://doi.org/10.18653/v1/D19-1608} {{Q}ua{RT}z: An open-domain dataset of qualitative relationship questions}.
\newblock In \emph{Proceedings of the 2019 Conference on Empirical Methods in Natural Language Processing and the 9th International Joint Conference on Natural Language Processing (EMNLP-IJCNLP)}, pages 5941--5946, Hong Kong, China. Association for Computational Linguistics.

\end{thebibliography}
\bibliographystyle{acl_natbib}

\appendix

\end{document}